\documentclass[journal]{IEEEtran}

\usepackage{url}  
\usepackage{graphicx} 
\usepackage{amsmath,amssymb}
\usepackage{color}
\usepackage{multirow}
\usepackage{bbm}
\usepackage{amsfonts}
\usepackage{booktabs}
\usepackage{bm}
\usepackage{cite}
\usepackage{ifpdf}
\usepackage{booktabs}

\usepackage[linesnumbered, boxed, ruled]{algorithm2e}

\usepackage[breaklinks=true,colorlinks,bookmarks=true]{hyperref}

\begin{document}

\title{Contrastive Transformer Learning with Proximity Data Generation for Text-Based Person Search}

\author{Hefeng Wu, Weifeng Chen, Zhibin Liu, Tianshui Chen, Zhiguang Chen, Liang Lin
\thanks{This work was supported in part by National Natural Science Foundation of China (NSFC) under Grant No. 62272494, 61876045, 62206060 and 61836012, Guangdong Basic and Applied Basic Research Foundation under Grant No. 2023A1515012845 and 2023A1515011374, and Fundamental Research Funds for the Central Universities, Sun Yat-sen University, under Grant No. 23ptpy111. (Corresponding author: Liang Lin)}
\thanks{Hefeng Wu, Weifeng Chen, Zhibin Liu, and Liang Lin are with the GuangDong Province Key Laboratory of Information Security Technology, School of Computer Science and Engineering, Sun Yat-sen University, Guangzhou, China (e-mail: wuhefeng@mail.sysu.edu.cn, chenwf35@mail2.sysu.edu.cn, liuzhb26@mail2.sysu.edu.cn, linliang@ieee.org).}%
\thanks{Zhiguang Chen is with the School of Computer Science and Engineering, Sun
Yat-sen University, Guangzhou, China, and also with the National Supercomputer Center in Guangzhou, Sun Yat-sen University, China (e-mail: zhiguang.chen@nscc-gz.cn).}%
\thanks{Tianshui Chen is with Guangdong University of Technology, Guangzhou, China (e-mail: tianshuichen@gmail.com).}
}
 
\markboth{IEEE Transactions on Circuits and Systems for Video Technology}%
{Wu \MakeLowercase{\textit{et al.}}: }

\maketitle

\begin{abstract}

Given a descriptive text query, text-based person search (TBPS) aims to retrieve the best-matched target person from an image gallery. Such a cross-modal retrieval task is quite challenging due to significant modality gap, fine-grained differences and insufficiency of annotated data. To better align the two modalities, most existing works focus on introducing sophisticated network structures and auxiliary tasks, which are complex and hard to implement. In this paper, we propose a simple yet effective dual Transformer model for text-based person search. By exploiting a hardness-aware contrastive learning strategy, our model achieves state-of-the-art performance without any special design for local feature alignment or side information. Moreover, we propose a proximity data generation (PDG) module to automatically produce more diverse data for cross-modal training. The PDG module first introduces an automatic generation algorithm based on a text-to-image diffusion model, which generates new text-image pair samples in the proximity space of original ones. Then it combines approximate text generation and feature-level mixup during training to further strengthen the data diversity. The PDG module can largely guarantee the reasonability of the generated samples that are directly used for training without any human inspection for noise rejection. It improves the performance of our model significantly, providing a feasible solution to the data insufficiency problem faced by such fine-grained visual-linguistic tasks. Extensive experiments on two popular datasets of the TBPS task (i.e., CUHK-PEDES and ICFG-PEDES) show that the proposed approach outperforms state-of-the-art approaches evidently, e.g., improving by 3.88\%, 4.02\%, 2.92\% in terms of Top1, Top5, Top10 on CUHK-PEDES.

\end{abstract}


\IEEEpeerreviewmaketitle


\section{Introduction}
\begin{figure}[!t]
   \centering
  \includegraphics[width=0.99\linewidth]{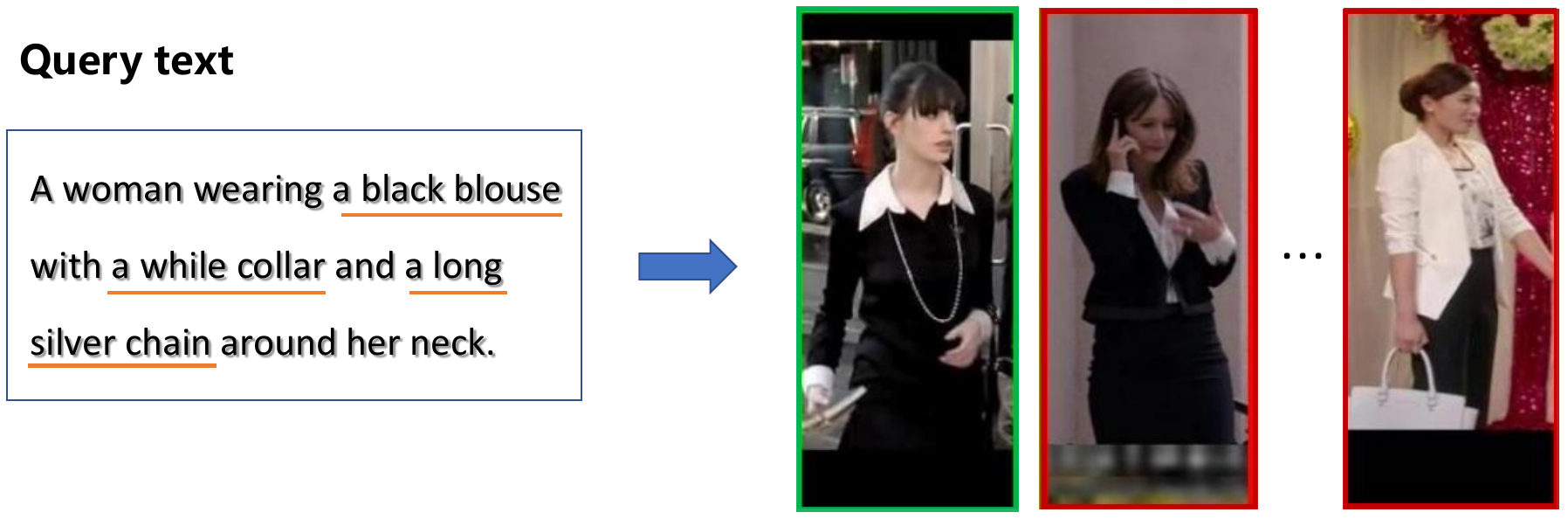}
   \caption{Illustration of text-based person search. In human-computer interaction, given a fine-grained natural language description of a person, the intelligent agent is asked to identify the targeted person in the image gallery. The green box denotes the correct person. }

   \label{fig:illustrate}
\end{figure}

\IEEEPARstart{P}{erson} search aims to retrieve the target person from an image gallery based on a given query, which plays a fundamental role in a large variety of real-world applications. While previous methods mostly focus on image-based person search (also known as person re-identification \cite{HaoLuo2019BagOT,SunZLYTW21pami}),  increasing attention has recently been drawn to text-based person search (TBPS) \cite{li2017person,farooq2021axm}, a more challenging cross-modal task in artificial intelligence (AI) that uses language descriptions only to retrieve the correct person, as illustrated in Figure~\ref{fig:illustrate}. It is more practicable and flexible for many real-world scenarios where an image of the target person is lacking.

However, text-based person search is far from being solved due to three main challenges: (i) \textbf{Cross modality:} the modalities of image and text are quite different. Image signal is continuous and redundant for semantic representation, while text information is discrete, whose semantic can be easily changed. (ii) \textbf{Fine-grained differences:} In contrast to generic image-text retrieval \cite{JiasenLu2019ViLBERTPT}, person search is a fine-grained task (refer to Figure~\ref{fig:illustrate}), whose description contains more detailed information about the only category, person. It is very challenging to distinguish the target person from other similar people (hard samples) based on a free-style sentence due to the subtle inter-class differences. (iii) \textbf{Data insufficiency:} Different from many other computer vision-related tasks, fine-grained language descriptions in this task are much more difficult and expensive to annotate. Thus, the performance of the model will be limited by the insufficient data of existing datasets. Meanwhile, since sentences may be ambiguous, the annotation is also prone to noise.

To align the two different modalities (i.e., images and texts), existing methods \cite{farooq2021axm,wang2020vitaa,aggarwal2020text} generally adopt convolutional neural networks (CNN) \cite{KaimingHe2016DeepRL,ChenLCHW22pami} to extract visual representation from images and use recurrent neural networks (RNN) \cite{SeppHochreiter1997LongSM} or Transformers \cite{AshishVaswani2017AttentionIA} to extract textual representations from sentences. Recently, Transformers have been demonstrated to perform well in various tasks (e.g., BERT \cite{devlin2018bert} and ViT \cite{dosovitskiy2021imageICLR}). Some works also use Transformers for the TBPS task. They commonly use BERT to extract textual presentation \cite{ChenZLWZ22Neurocompt,gao2021contextual}. 
These methods still combine with CNNs for visual representations to form complicated architectures. 
Using Transformers for both textual and visual representations is not well explored for this task. One significant reason may be that although Transformer structures are more powerful than CNNs, they are also more data-hungry, so it is more challenging to align the two modalities with only Transformers for the TBPS task that has insufficient training data.
In contrast to previous works, in this work we investigate a dual Transformer model, which is simple but effective, to learn the common semantic representation of images and texts for this task. We demonstrate that with proper learning algorithms, our dual Transformer model can outperform all previous approaches.

In order to learn fine-grained cross-modal features, most existing methods either design multi-branch networks \cite{niu2020improving} to extract both local and global features from images and match them with phrase-level and sentence-level text features, or utilize extra auxiliary tasks like pose estimation \cite{jing2020pose} and semantic segmentation \cite{wang2020vitaa} to help with the modality alignment. Although these sophisticated strategies are proven useful for fine-grained matching, they are complex and hard to implement in practice. In this paper, we simply introduce a hardness-aware contrastive learning strategy, which can help our model learn to distinguish hard samples and achieve better performance. It is worth noting that, without applying any additional side information or local alignment, our plain model can learn better fine-grained semantic representation and achieve good cross-modal alignment implicitly.

Moreover, considering the great scale difference between the huge model parameters and the limited data of existing TBPS datasets, we design a proximity data generation (PDG) module to overcome the overfitting issue on the performance of our Transformer model due to data insufficiency. Inspired by the recent cross-modal generative AI models \cite{SahariaCSLWDGLA22nips,Ramesh2022ArXivHierarchical,ChenWXWLXL2023ArXiv}, the PDG module develops an automatic algorithm based on the leading text-to-image 
diffusion model to generate controlled text-image pair samples effectively in the proximity space of original ones, and it further incorporates conservative approximate text generation and feature-level mixup to strengthen data diversity. The PDG module effectively enriches the training data and helps the model learn better decision boundary in contrastive learning. To our best knowledge, we are the first to explore such algorithm for this task. We demonstrate that the proposed algorithm can bring considerable improvement to our model.

In summary, our work makes the following contributions. (1) We introduce a simple dual Transformer model equipped with a hardness-aware contrastive learning strategy for text-based person search.
Without any sophisticated architecture designs, our simple model can achieve better performance than previous leading methods. To our best knowledge, we are the first to introduce such Transformer model and make it work well for the TBPS task that is limited by insufficient data. 
(2) We propose a proximity data generation module to enrich the training data and help learning better discriminative cross-modal representation to improve model performance, where such algorithm is firstly explored for the TBPS task. It provides a feasible solution to address the data insufficiency problem of such fine-grained visual-linguistic tasks when the annotation is difficult and expensive.  
(3) We conduct extensive experiments on two widely-used TBPS datasets, i.e., CUHK-PEDES \cite{li2017person} and ICFG-PEDES \cite{ding2021semantically}, and experiments show that our method outperforms state-of-the-art methods evidently, e.g., 3.88\%, 4.02\%, 2.92\% improvements in terms of Top1, Top5, Top10 accuracy for CUHK-PEDES, and 0.27\%, 0.82\%, 1.22\% for ICFG-PEDES.

\begin{figure*}[!t]
   \centering
   \includegraphics[width=0.96\linewidth]{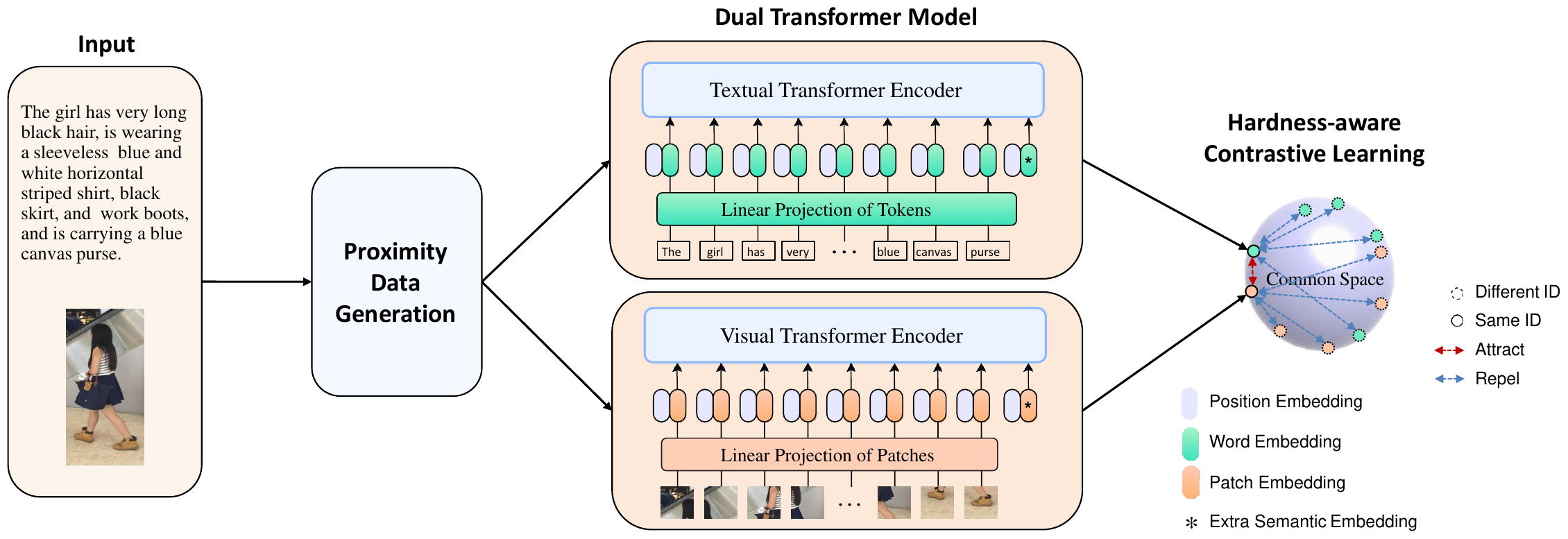}
   \caption{Illustration of our learning framework for text-based person search. The dual Transformer model handles both textual and visual modalities.  The proximity data generation module produces proximate data to address the data insufficiency problem of this task. It facilitates learning better decision boundary in hardness-aware contrastive learning that is designed to be sensitive to hard negative pairs and push them away.} \label{fig:framework}  
\end{figure*}

\section{Related Work}
\subsection{Text-Based Person Search}
\label{relate_work}
Text-based person search is first proposed by \cite{li2017person}, and they utilize a CNN model for images and a RNN model with gated attention for texts to address this task. Zhang and Lu \cite{Zhang_2018_ECCV} propose a new loss function that uses the KL divergence for two modality matching, which is also adopted by many works.  Zheng et al. \cite{ZhedongZheng2020DualpathCI} propose a CNN+CNN structure with a ranking loss. However, CNN fails to capture long-term relations, which is not suitable for text feature extraction. Niu et al. \cite{niu2020improving}  try to align the features from two modalities at the levels of global-to-global, global-to-local, and local-to-local by using cross-modal attention, which is inefficient since each image-text pair needs plenty of computation to get the similarity at testing.
Besides, some methods utilize side information to better extract the visual features. Jing et al. \cite{jing2020pose} use pose information to guide the network to learn fine-grained alignment (noun phrases and human pose region alignment). Aggarwal et al. \cite{aggarwal2020text} propose to utilize extra attribute labels to make up for the lack of text information. Moreover, Wang et al. \cite{wang2020vitaa} use a semantic segmentation network to get each part of person, and use these features to guide the network to learn local-to-local features. Recently, Gao et al. \cite{gao2021contextual} propose a staircase CNN network and a local constrained BERT model to align local and global features from two modalities. Chen et al. \cite{ChenZLWZ22Neurocompt} use a multi-branch CNN to extract image features from local and global perspectives and use BERT+CNN to extract text features.
Li et al. \cite{LiHui2022Transformer} put visual and textual features into the same Transformer to predict the matching result.
Shao et al. \cite{ShaoZFLWD22mm} use BERT for textual representation and learn granularity-unified representations for both modalities to promote TBPS performance.
Suo et al. \cite{SuoSNGWZW22eccv} use BERT and ResNet models for text and image representations respectively, and introduce lightweight correlation filters for cross-modal alignment in global and local levels.
To achieve a better performance, these methods generally exploit sophisticated designs to incorporate side information or extract multi-scale features. In contrast, we introduce a simple framework that can be implemented easily using existing Transformer models out of the box, and it outperforms all current leading methods.

\subsection{Transformer-Based Language and Vision Models}

Transformer is first proposed by \cite{AshishVaswani2017AttentionIA} for sequential tasks like machine translation and has drawn a lot of attention since BERT \cite{devlin2018bert} is proposed and demonstrated the ability to handle most natural language processing tasks. Recently, ViT \cite{dosovitskiy2021imageICLR} successfully adapts Transformer to computer vision and achieves state-of-the-art performance by training on large-scale data. Since Transformer is more data-hungry due to lack of inductive bias, BeiT \cite{HugoTouvron2020TrainingDI} proposes a distillation framework to train the ViT without the need of large-scale datasets.
Transformer-based models are also implemented in the field of generic cross-modal tasks. There are two paradigms of models. Single-stream models like Oscar \cite{XiujunLi2020OscarOA} concatenate two modalities of input and forward them through Transformer, which can achieve better performance at the expense of inference speed. Two-stream models like ViLBERT \cite{JiasenLu2019ViLBERTPT} use two separate models to extra semantic representation from both modalities, which is more efficient since each image or text can get their representation via one-time calculation. 
Radford et al. \cite{RadfordKHRGASAM21icml} propose the contrastive language-image pre-training (CLIP) model that learns from scratch on a dataset of 400 million image-text pairs collected from the internet, demonstrating good zero-shot transfer ability to downstream tasks. 
Following that, FLIP \cite{YaoHHLNXLLJX22iclr} uses a cross-modal late interaction mechanism to achieve finer-level alignment between modalities.
BLIP \cite{LiXH22icmlBLIP} exploits bootstrapping in training and unifies vision-language understanding and generation tasks in one learning framework. 
However, These methods rely on large-scale datasets and the research of Transformer-based models for small datasets in cross-modal fine-grained tasks is still rare.
In this paper, we propose a simple model for text-to-image person search, which utilizes the powerful Transformer architecture to achieve state-of-the-art performance.

\subsection{Text-to-Image Generation}
Recently, great success has been witnessed in generative AI models, bringing widespread and profound impact. 
Among them, plenty of works \cite{BrockDS19iclr,KarrasLA21pami,SahariaCSLWDGLA22nips,Ramesh2022ArXivHierarchical} have been proposed to address the text-to-image generation task that generates images conditioned on plain text.  
Compared with generative adversarial networks (GANs) \cite{BrockDS19iclr,KarrasLA21pami,KarrasALHHLA21nips,LiYXM23pami}, the recent diffusion models are witnessed to achieve more promising generation results. Several large-scale diffusion models, such as Imagen \cite{SahariaCSLWDGLA22nips}, DALL-E 2 \cite{Ramesh2022ArXivHierarchical}, and the open-source Stable Diffusion \cite{RombachBLEO22cvpr}, demonstrate impressive generation quality. Since harvesting an effective diffusion model from scratch requires massive and time-consuming training on billions of data, many works \cite{Ruiz2023DreamBooth} are devoted to the customization and control of the pretrained diffusion models.
DreamBooth \cite{Ruiz2023DreamBooth} fine-tunes text-to-image diffusion models for subject-driven generation that takes several images of a subject as input and generates images of the subject in different contexts with text prompt guidance. 
Hertz et al. \cite{Hertz2023Prompt2PromptICLR} present a prompt-to-prompt image editing method that modifies some words of a text prompt to change certain parts of the generated image by altering the semantically related cross-attention maps in the model.
DiffEdit \cite{Couairon2022DiffEdit} automatically generates a mask to  highlight regions of the input image that need to be edited and preserves other image contents with mask-based diffusion. 
Mokady et al. \cite{Mokady2023Nulltext} introduces a null-text inversion method that enables the diffusion model to generate the given real image with high fidelity. Inspired by these works, we design an automatic algorithm pipeline to generate proper text-image pair samples of to help address the fine-grained TBPS task. 

\section{Methodology}\label{sec:Methodology}
In this section, we describe the proposed method in detail. Figure~\ref{fig:framework} illustrates our overall learning framework. We will present the dual Transformer model in Section \ref{sec:model}, followed by the proximity data generation module in Section \ref{sec:dataGeneration}. Afterwards, we describe the hardness-aware contrastive learning strategy and the training loss in Section \ref{sec:loss}. For testing, only the dual Transformer model is needed for inference.

\subsection{Dual Transformer Model}
\label{sec:model}
Our dual Transformer model is illustrated in Figure \ref{fig:framework}. It consists of a visual Transformer for image encoding and a textual Transformer for text encoding. In our model design, the two Transformer encoders follow the original Transformer structure \cite{AshishVaswani2017AttentionIA} closely so that we can use existing efficient Transformer implementations out of the box.
Both two Transformer encoders have $L$ Transformer layers. Each layer contains multi-head self-attention (MSA) and multi-layer perceptron (MLP) blocks, which can be formulated as:
\begin{align}
\mathbf{z}'_{l}&=MSA(\mathbf{z}_{l-1})+\mathbf{z}_{l-1}\,, & l=1,...,L\\
\mathbf{z}_{l}&=MLP(\mathbf{z}'_{l})+\mathbf{z}'_{l}\,, &l=1,...,L
\end{align}
where $\mathbf{z}_{l}$ is the sequence of encoded tokens output at layer $l$ and layer normalization is applied for each block. $\mathbf{z}_{0}$ is the input sequence of embedded tokens, denoted as:
\begin{equation}
\mathbf{z}_{0}=[\mathbf{E}_{sem}+\mathbf{E}_{pos}^0;\mathbf{E}_{1}+\mathbf{E}_{pos}^1;\ldots;\mathbf{E}_{N}+\mathbf{E}_{pos}^N],
\end{equation}
where $N$ is the number of tokens, $\mathbf{E}_{k}\in \mathbb{R}^{D}$ is an embedding vector of size $D$, and $\mathbf{E}_{pos}^k\in \mathbb{R}^{D}$ is the position embedding. Similar to BERT's [class] token, we add an extra learnable token $\mathbf{E}_{sem}$, whose corresponding state $\mathbf{z}_{L}^0$ at layer $L$ serves the semantic representation of the input text or image.

\textbf{Textual Representation Extraction:} The textual Transformer takes sentences as input. The input sentence will first be tokenized and then projected to one-dimensional word embedding via a trainable linear projection to obtain $\mathbf{z}_{0}$. The output $\mathbf{z}_{L}^0$ of the last layer $L$ is the semantic representation of the input sentence.

\textbf{Visual Representation Extraction:} The visual Transformer takes images as input. We follow ViT \cite{dosovitskiy2021imageICLR} to turn an input image into a sequence of tokens. Given an input image $x\in R^{H\times W\times C}$, where $H$, $W$, $C$ denote the height, width and channel number, respectively, we split it into $N_p$ fixed-sized two-dimensional patches $x_p^k \in \mathbb{R}^{P^2\times C}$, $k=1,...,N_p$, where $P$ is the patch size, and the number of patches $N_p=H$ $\times$ $W/P^2$. Then we flatten the patches and project them to one-dimensional visual embedding via a linear projection. Likewise, the output $\mathbf{z}_{L}^0$ of the last layer $L$ is the semantic representation of the input image.

\textbf{Overlapping Slicing:} Since non-overlapping slicing in original ViT may ignore some distinguished information of the input image, we propose to use overlapping slicing for image tokens split. We denote the stride step as $w$, and then the number of patches is:
\begin{equation}
N_p = \frac{(H+w-P)\times (W+w-P)}{w^2}.
\end{equation}

Compared to the non-overlapping one, overlapping slicing can extract information from neighbour patches and enhance the representation learning of the visual encoder.

\textbf{Cross-Modal Matching}: %
As aforementioned, the aggregate feature representation of texts and images will be output by the symmetric Transformer encoders, respectively. Given a text description $s$ and an image $x$, let $\mathbf{f}_{T}$ and $\mathbf{f}_{I}$ denote the output textual and visual representation. We measure the similarity between the two modalities using the cosine similarity metric, which can be formulated as:
\begin{equation}
\begin{aligned}
S = \text{Sim}(s, x) = \frac{\mathbf{f}_{T}\cdot \mathbf{f}_{I}}{||\mathbf{f}_{T}||\times ||\mathbf{f}_{I}||} \,, 
\end{aligned}
\end{equation}
where $||\mathbf{f}||$ denotes the $L_2$ norm of $\mathbf{f}$, and the operator $\cdot$ is the dot product of vectors. Note that both $\mathbf{f}_{T}$ and $\mathbf{f}_{I}$ have the same dimension $D$, and each represents a point in a common hyper-sphere. The training goal is to maximize the similarity of positive pairs (the image and its corresponding caption) and minimize the similarity of negative pairs (the image and any other caption in the batch).

\subsection{Proximity Data Generation}
\label{sec:dataGeneration}
So far, text-to-image person search is faced with a challenge that is seldom addressed by existing methods, i.e., the lack of sufficient training data. Existing datasets of this task are orders of magnitude smaller than those generic image-text pair datasets \cite{ChenLYK0G0020eccv,Li0LZHZWH0WCG20eccv}, because it is much more difficult and expensive to annotate a large-scale dataset with highly fine-grained text descriptions.
Moreover, our model is based on Transformers, which lacks of inductive bias and is more data-hungry, making it overfit on small datasets more easily.

Therefore, in this work we design a proximity data generation module to promote our model's performance by explicitly addressing the data insufficiency problem. To our best knowledge, we are the first to exploit such strategies for this task. 

Our proximity data generation module includes three parts to generate training data from different perspectives: (i) Controlled text-image pair generation, which utilizes leading text-to-image diffusion models to generate new person images corresponding to the modified text descriptions and thus provides new text-image pairs for training; (ii) Approximate text generation, which generates a new approximate text description for a given text-image pair and replaces the original text; and (iii) Feature-level mixup, which mixes up the features of two text-image pairs to generate a new text-image feature pair. 
We explore these new strategies to generate diversified and effective examples in the proximity space of original examples. As verified in our experiments, these examples can improve the model performance effectively. 
In the following, we will describe them in detail.

\begin{figure}[!th]
   \centering
  \includegraphics[width=0.92\linewidth]{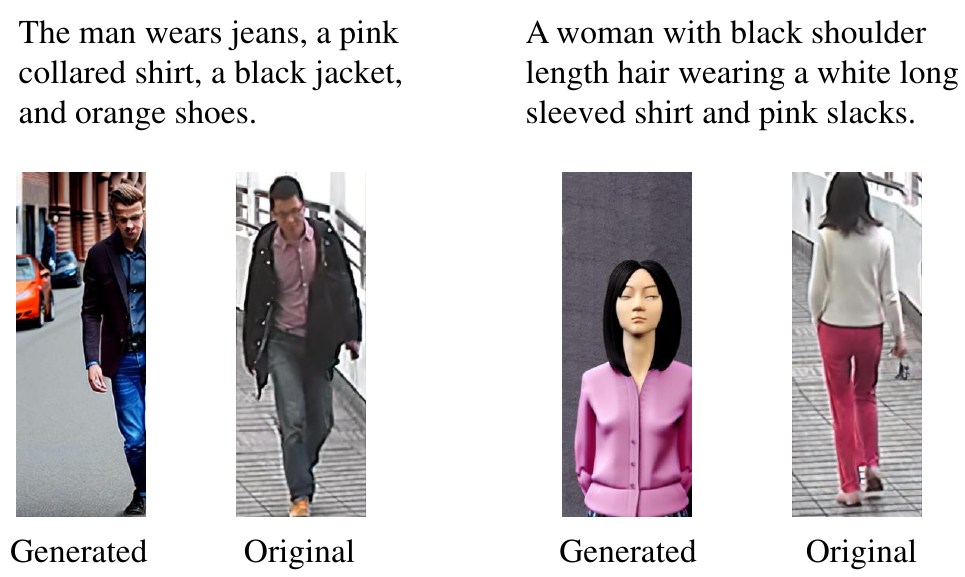}
   \caption{Images directly generated by the Stable Diffusion model with fine-grained natural language description as input. The original person images are exhibited on right for comparison. It can be observed that the details of the generated person images are not matched with the text.}
   \label{fig:oriGen}
\end{figure}

\begin{figure*}[!t]
   \centering
  \includegraphics[width=0.8\linewidth]{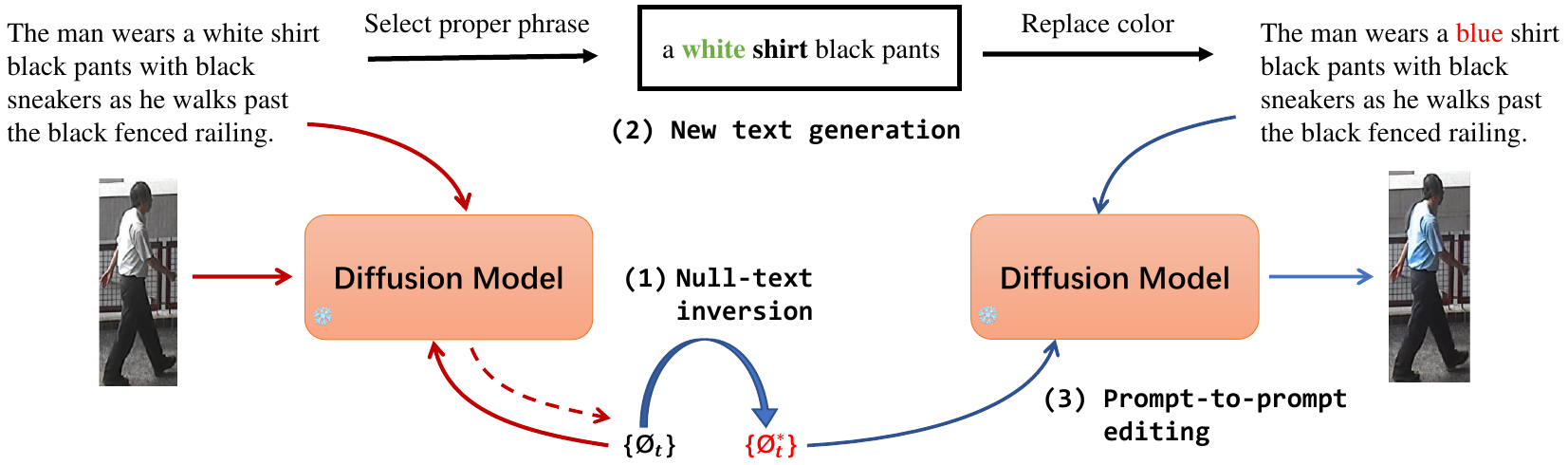}
   \caption{Illustration of the automatic pipeline of our controlled text-image pair generation. (1) Given the original text-image pair, we utilize the null-text inversion method to generate optimized null-text embeddings which can make the Stable Diffusion model reconstruct the given person image when using the given text as input. (2) We select a proper phrase  and replace the color word randomly to generate a new text. (3) We use the prompt-to-prompt editing method to generate a new person image.}
   \label{fig:T2Igen}
\end{figure*}

\textbf{Controlled Text-Image Pair Generation:} 
Recently, text-to-image generation based on diffusion models has made great progress, which inspires us to use such technique to generate new text-image pairs of persons and add to the training data. 

We first describe the diffusion model \cite{HoJA20nips,LugmayrDRYTG22cvpr,Ramesh2022ArXivHierarchical} briefly.
Given an input signal $x_0$, a diffusion forward process is adding noise to $x_0$ step by step and resulting in a noise $x_K$ that nearly follows an isotropic Gaussian distribution, where $K$ is the total timestep.
The diffusion model aims to learn to reverse the diffusion process (denoising). Given a random noise $x_K$,  the diffusion model with parameters $\theta$ will predict the noise added at the previous timestep $x_{t-1}$ until the original signal $x_0$ is recovered. The reverse Gaussian transition is given by  
\begin{equation}
    p_\theta(x_{t-1} | x_t) = \mathcal{N}(x_{t-1}; \mu_\theta(x_t,t), \Sigma_\theta(x_t,t) ), ~t = K, ..., 1,
\end{equation}
where $\mu_\theta(x_t,t)$ denotes the mean and $\Sigma_\theta(x_t,t)$ denotes the variance. By following the denoising
diffusion implicit models (DDIM) formulation \cite{SongME21iclr}, we have 
\begin{equation}
    x_{t-1} = \alpha_{1,t} \cdot x_t + \alpha_{2,t}  \cdot \varepsilon_{\theta}(x_t, t, \mathcal{C}),
\end{equation}
where $\varepsilon_{\theta}(x_t, t, \mathcal{C})$ is the predicted noise at timestamp $t$ and $\mathcal{C}$ is the conditional text description for guidance. $\alpha_{1,t}$ and $\alpha_{2,t}$ are coefficients calculated at timestamp $t$. Please refer to \cite{SongME21iclr,Mokady2023Nulltext} for more details.

In this work, we utilize the publicly available Stable Diffusion model \cite{RombachBLEO22cvpr} as our basis. However, though the Stable Diffusion model has been pretrained on billions of general text-image pairs, directly applying it cannot obtain acceptable examples for this task. As exhibited in Figure \ref{fig:oriGen}, we take  text-image pair samples from existing TBPS datasets and feed the text descriptions into the Stable Diffusion model to generate new person images and compare with the original images. It can easily be observed that the fine-grained details of person are not matched with the text. Even sometimes a random noise results in a properly-matched person image, we have to pick it out manually, which is far away from our goal of an automatic and valid generation pipeline. 

Therefore, we design a new automatic pipeline that generates new text-image samples in the proximity space of original text-image person pairs.
To better achieve this, we first use the training samples in the given  TBPS dataset to fine-tune the Stable Diffusion model so that it can generate person images that better adapt to the target domain. The model is optimized by minimizing a simple noise-prediction loss:
\begin{equation}
min_\theta || \varepsilon - \varepsilon_{\theta}(x_t, t, s) ||_2^2.
\end{equation}
where $x$ and $s$ is a text-image pair of person and $x_0=x$.

Based on the fine-tuned Stable Diffusion model, given a text-image pair $(s, x)$, we first employ the null-text inversion method \cite{Mokady2023Nulltext} to obtain a noise vector $x_K$ and a list of optimized null-text embeddings $\{\Phi_t^*\}_{t=1}^{K}$, which can be used with the diffusion model to accurately generate the person image $x$ when using $s$ as the input. Then we change specific words in the text $s$ and generate a new person image from the Stable Diffusion model by combining the prompt-to-prompt image editing method \cite{Hertz2023Prompt2PromptICLR}, which tries to modify the generated image reasonably by modifying the cross-attention maps during the diffusion process to adapt to the changed words. The whole pipeline is illustrated in Figure \ref{fig:T2Igen}.

In the above process, we generate a new text by replacing a word in the text $s$. We first use spaCy \cite{Honnibal2020spaCy}, a natural language processing toolbox, with its pretrained model ``en\_core\_web\_sm'' to divide text $s$ into phrases. Then we randomly select one phrase, in which the old color word is replaced with a new color, e.g., changing ``red'' to ``blue''. Actually, we have also tried to replace the clothes (e.g., changing ``shirt'' to ''coat'') in experiments, but the results are uncontrollable and unsatisfying. Moreover, we find that changing the color of small belongings (e.g., bags and backpacks) is also unsatisfying. Therefore, to make the generated image in proper control, we define a clothes set $Cl$ and a color set $Co$. A phrase with its clothes word appearing in $Cl$ is randomly selected and then the color word is replaced with a new color randomly selected from $Co$. 

\begin{figure*}[!t]
   \centering
  \includegraphics[width=0.93\linewidth]{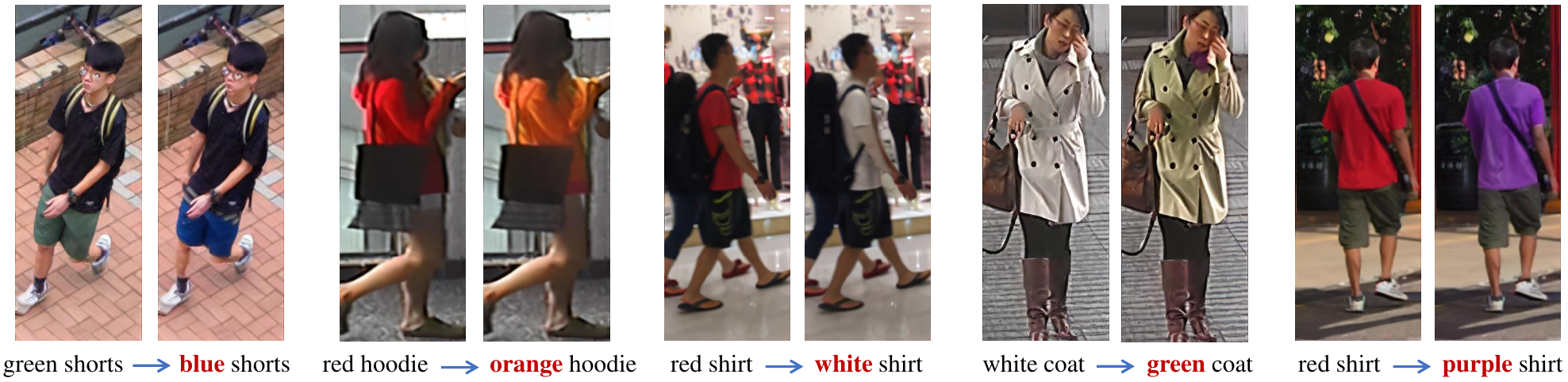}
   \caption{Example person images generated by our automatic pipeline. The left is the original image and the right is the generated image. The randomly selected phrase is shown below and the new color is highlighted in red.}

   \label{fig:genExamples}
\end{figure*}

In summary, the whole algorithm is outlined in \textbf{Algorithm} \ref{alg:T2Igen}. Its pipeline is fully automatic and every sample it generates is acceptable for use. Some generated image examples are exhibited in Figure \ref{fig:genExamples}.

\begin{algorithm}[h]
\SetAlgoLined
 \caption{Controlled text-image pair generation}\label{alg:T2Igen}
\textbf{Input:} A text-image pair $(s, x)$\\ 
\textbf{Output:} A new text $s'$ and a new generated image $x'$\\
 \vspace{1mm} \hrule \vspace{1mm}
 Input $(s, x)$ to the null-text inversion model and obtain noise $x_K$ and null-text embeddings $\{\Phi_t^*\}_{t=1}^{K}$; \\
 Split $s$ into phases with spaCy; \\
 Randomly select a phase whose clothes word appears in set $Cl$; \\
 Replace the color word in the selected phase by a new color chosen in set $Co$ and form the new text $s'$;\\
 Input $x_K$, $\{\Phi_t^*\}_{t=1}^{K}$, and $s'$ to the diffusion model and generate a new person image $x'$ by combining the prompt-to-prompt editing method; \\
 \textbf{Return} $s'$, $x'$
\end{algorithm}

\textbf{Approximate Text Generation:} Given a person image $x$ and its text description $s$, we utilize the following operations to generate approximate text $s'$:
\begin{equation}\label{eq:textOP}
 s' = \mathcal{F}_{*}(s,\sigma), \quad * \in \mathcal{D}_{T}=\{ \text{SDEL}, \text{CDEL}, \text{REPL}\}
\end{equation}
where $\sigma \in (0,1)$ is a parameter denotes the portion of words being altered for $s$, and $\mathcal{F}_{*}$ will generate approximate sentences by altering $N_w$ words:
\begin{equation}\label{eq:}
 N_w = \lfloor \mathcal{W}(s)\cdot \sigma \rfloor,
\end{equation}
where $\mathcal{W}(s)$ denotes the number of words in the sentence $s$, and $\lfloor u \rfloor$ denotes the largest integer not greater than $u$.

Since altering a sentence can easily change its semantics, the three operations in $\mathcal{D}_{T}$ are conservative.
To be specific, $\mathcal{F}_{\text{SDEL}}$ will randomly select $N_w$ single words of the text to delete, $\mathcal{F}_{\text{SDEL}}$ will randomly select $N_w$ continuous words of the text to delete, and $\mathcal{F}_{\text{SDEL}}$ will randomly select $N_w$ single words of the text and replace each selected word with
a synonym according to the WordNet \cite{GeorgeAMiller1995WordNetAL}.
By setting the parameter $\sigma$ to a small value, our operations have a low risk of changing the semantics badly. In our experiments, $\sigma$ is empirically set as 0.2.

In training, for an image $x$ and its description $s$, we set a probability of 0.5 to generate the approximate text $s'$ for $s$, using one of the three operations randomly. Once $s'$ is generated, we enforce the following constraint in optimization:
\begin{equation}\label{eq:constraint}
 \text{Sim}(s, x)>\text{Sim}(s', x)>\text{Sim}(s^-, x),
\end{equation}
where $s^-$ denotes any text description for other images. In this way, we can use $s'$ to enrich the training data and further push away $s^-$ and $x$.

\textbf{Feature-level Mixup:} Given two positive text-image pairs, i.e., $s_1$ and $x_1$, $s_2$ and $x_2$, we further propose feature-level mixup to generate a new pair for cross-modal matching. Specifically, the mixup is performed on the hidden states output from the first Transformer layer, formulated as:
\begin{align}
\mathbf{\hat{z}}^T_{1} &= \lambda \mathbf{z}^T_{1,1} + (1-\lambda)\mathbf{z}^T_{1,2}, \quad \mathbf{z}^T_{1,1},~\mathbf{z}^T_{1,2}~\text{for}~ s_1,~s_2\\
\mathbf{\hat{z}}^I_{1} &= \lambda \mathbf{z}^I_{1,1} + (1-\lambda)\mathbf{z}^I_{1,2}, \quad \mathbf{z}^I_{1,1},~\mathbf{z}^I_{1,2}~\text{for}~ x_1,~x_2
\end{align}
where $\lambda$ is a parameter controlling the mixup degree. 
Then we separately forward the generated hidden-state features $\mathbf{\hat{z}}^T_{1}$ and $\mathbf{\hat{z}}^I_{1}$ through the rest Transformer layers to get the final mixup representation, and treat them as a new positive pair. 

Feature-level mixup enriches the data by generating new pairs in the proximity feature space of original ones. For the cross-modal TBPS task, we propose to mix the data from both modalities at the feature level, because interpolating the features of hidden states  does weaker harm to the semantics of the sentence.

\begin{figure}[!t]
   \centering
  \includegraphics[width=0.99\linewidth]{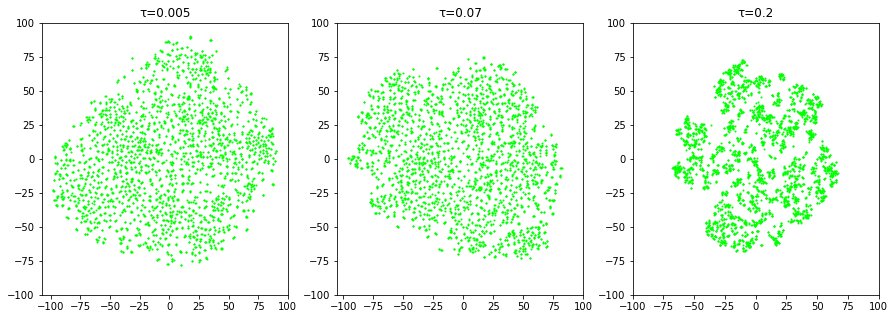}
   \caption{T-SNE visualization of image embedding distribution of the test set. As shown in the figure, smaller temperature tends to generate more uniform distribution so that similar identities can be better distinguished.}
   \label{fig:img_embed_distribution}
\end{figure}

\subsection{Hardness-Aware Contrastive Learning}
\label{sec:loss}
To enable our model to learn a common semantic representation from both images and texts, we introduce a contrastive learning strategy to minimize the distance of positive image-text pairs and maximize the distance of negative pairs, whose loss function stems from the InfoNCE Loss \cite{oord2018representation}. 
We utilize this contrastive loss for our fine-grained cross-modal person retrieval task.

Given a batch of $N$ image-text pairs, each pair takes text embedding $\mathbf{f}_{T}$ as query and its corresponding image embedding $\mathbf{f}_{I}$ as a positive key while other image embeddings in the batch as negative ones. The text-to-image loss function for pair $j$ can be formulated as:
\begin{equation}\label{eq:t2i}
\begin{aligned}
L_{t2i}(j) = -\log{\frac{\exp({\mathbf{f}_{T}^j\cdot \mathbf{f}_{I}^{j}/\tau})}{\sum_{k=1}^{N} \exp({\mathbf{f}_{T}^j\cdot \mathbf{f}_{I}^k/\tau})}}.
\end{aligned}
\end{equation}
When query $\mathbf{f}_{T}^j$ is similar to its positive key $\mathbf{f}_{I}^{j}$ and dissimilar to all other keys in the batch, the value of loss $L_{t2i}(j)$ is low. In other words, the loss function tries to pull the positive pairs close  and push the negative pairs away. Since our text-to-image contrastive loss is asymmetric for each input modality, we also define an image-to-text loss:
\begin{equation}\label{eq:i2t}
\begin{aligned}
L_{i2t}(j) = -\log{\frac{\exp({\mathbf{f}_{I}^j\cdot \mathbf{f}_{T}^{j}/\tau})}{\sum_{k=1}^{N} \exp({\mathbf{f}_{I}^j\cdot \mathbf{f}_{T}^k/\tau})}} .
\end{aligned}
\end{equation}

When the approximate text $s'_j$ is generated for $s_j$, $s'_j$ will replace $s_j$ in Eqs. (\ref{eq:t2i}) and (\ref{eq:i2t}), and a regularization term is defined to ensure the constraint (\ref{eq:constraint}):
\begin{equation}\label{eq:}
L_{c}(j) = \max(0,\frac{\mathbf{f}_{I}^j\cdot \mathbf{f}_{T}^{j}}{\mathbf{f}_{I}^j\cdot \mathbf{\hat{f}}_{T}^{j}}-1),
\end{equation}
where $\mathbf{\hat{f}}_{T}^{j}$ is the embedding feature of the original text $s_j$.

The total loss of a batch is the sum of the losses of all pairs, formulated as:
\begin{equation}
\begin{aligned}
\mathcal{L} = \sum_{j=1}^N L_{t2i}(j) + L_{i2t}(j) + \mu\Phi(j)L_{c}(j) ,
\end{aligned}
\end{equation}
where $\mu$ is a balancing parameter, and $\Phi(j)$ is an indicator that takes 1 if $s_j$ is replaced by the approximate text $s'_j$ and 0 otherwise.

Note that, in this fine-grained TBPS task, we will show that the performance of our model is sensitive to the temperature parameter $\tau$, which plays a role in controlling the strength of penalties of hard negative samples \cite{wang2021understanding}. 
For illustration, we plot t-SNE visualization of the image embedding distribution of the test set in CUHK-PEHDES with respect to different values of $\tau$. As exhibited in Figure \ref{fig:img_embed_distribution}, smaller temperature tends to generate more uniform distribution, which is helpful for distinguishing hard samples in our fine-grained task.
A more theoretical explanation is given below. Let $S_+ = \mathbf{f}_{T} \cdot \mathbf{f}_{I}^+$ and $S_- = \mathbf{f}_{T} \cdot \mathbf{f}_{I}^-$, which represent similarities of a positive pair and a negative pair, respectively. Thus, we can define the relative penalty on one negative sample $j$ with respective to the positive sample:
\begin{equation}
\begin{aligned}
r_j = |\frac{\partial L_{t2i}}{S_{j-}}| / |\frac{\partial L_{t2i}}{S_+}| = \frac{\exp({S_{j-}/\tau})}{\sum_{k=1}^{N-1} \exp({S_{k-}/\tau})},
\end{aligned}
\end{equation}
where $k$ denotes the index of all $(N-1)$ negative samples and $\sum_{j=1}^{N-1} r_j = 1$. As the $\tau$ decreases, the distribution of $r_j$ becomes more sharp in large similarity region, which results in large penalty to the samples whose similarity is close to the positive pair.
Specifically, when $\tau \to 0$, the loss only focuses on the hardest (the most similar) negative sample while ignoring other negative samples, which is like the triplet loss that samples a positive pair and a hardest negative sample. With lower temperature parameter, the loss function can automatically mine the hard samples and give them a large penalty.

For a fine-grained task like person search, there are a large number of hard samples needed to be distinguished, since different people may have similar appearances. Thus, a smaller temperature can help the model focus on hard samples and learn stronger representations (as shown in Table \ref{tab:tau}).

\section{Experiments}
\subsection{Datasets and Evaluation Metrics}
We conduct extensive experimental evaluation and comparison on two text-based person search datasets, CUHK-PEHDES \cite{li2017person} and ICFG-PEDES \cite{ding2021semantically}. \\
\textbf{CUHK-PEHDES} has 40,206 images and a total of 80,412 sentences for 13,003 identities with 2 captions per image. The average word length is 23.5. The training set contains 34,054 images of 11,003 persons, while the validation and test sets have 3,078 images of 1,000 persons, 3,074 images of 1,000 persons, respectively. \\
\textbf{ICFG-PEDES} is a dataset with fewer identities, which contains 54,522 images of 4,102 identities with 1 caption per image. The average word length of sentence is 37.2. The training set contains 34,674 images of 3,102 persons, and the test set comprises 19,848 images of 1, 000 persons.

\noindent\textbf{Evaluation Metrics:} We follow most previous works to use the Top-k (k=1, 5, 10) accuracy as evaluation metrics. A successful search means that a matched person image exists within the Top-k retrieved images. Besides Top-k accuracy, we also adopt mean Average Precision (mAP) for evaluation. Empirically, Top-k reflects the model’s accuracy on the first few retrieval results while mAP focuses on the order of the entire retrieval images.

\begin{table}[!t]
\centering
\caption{Comparison results (\%) on CUHK-PEDES} 
\label{tab:cuhk_sota}
\begin{tabular}{|l|cccc|}
\hline
Method & Top1 & Top5 & Top10 & mAP \\
\hline
\hline
GNA-RNN \cite{li2017person} & 19.05 & - & 53.64 & - \\
DP-CNN \cite{ZhedongZheng2020DualpathCI} & 44.40 & 66.26 & 75.07 & - \\
CMPM \cite{Zhang_2018_ECCV} & 49.37 & 71.69 & 79.27 & - \\
MIA \cite{niu2020improving}  & 53.10 & 75.00 & 82.90 & - \\
PMA \cite{jing2020pose} & 54.12 & 75.45 & 82.97 & - \\
ViTAA \cite{wang2020vitaa}  & 54.92 & 75.18 & 82.90 & 51.60 \\
CMAAM \cite{aggarwal2020text} & 56.68 & 77.18 & 84.86 & - \\
NAFS \cite{gao2021contextual} & 59.36 & 79.13 & 86.00 & 54.07 \\
AXM-Net \cite{farooq2021axm} & 61.90 & 79.41 & 85.75 & 57.38 \\
SSAN \cite{ding2021semantically} & 61.37 & 80.15 & 86.73 & - \\
TIPCB \cite{ChenZLWZ22Neurocompt} & 63.63 & 82.82 & 89.01 & 56.78 \\ 
TextReID \cite{HanHZX21bmvc} & 64.08 & 81.73 & 88.19 & 60.08 \\
LGUR \cite{ShaoZFLWD22mm} & 64.21 & 81.94 & 87.93 & - \\
IVT \cite{ShuWWCSQRW22eccv} & 65.59 & 83.11 & 89.21 & - \\
\hline
\textbf{Ours} & \textbf{69.47} & \textbf{87.13} & \textbf{92.13} & \textbf{60.56} \\
\hline
\end{tabular}
\end{table}

\begin{table}[!t]
\centering
\caption{Comparison results (\%) on ICFG-PEDES}
\label{tab:icfg_sota}
\begin{tabular}{|l|cccc|}
\hline
Method & Top1 & Top5 & Top10  & mAP  \\ 
\hline
\hline
DP-CNN \cite{ZhedongZheng2020DualpathCI} & 38.99 & 59.44 & 68.41 & -  \\
CMPM \cite{Zhang_2018_ECCV} & 43.51 & 64.55 & 74.26 & -  \\
MIA \cite{niu2020improving}  & 46.49 & 67.14 & 75.18 & -  \\
ViTAA \cite{wang2020vitaa}  & 50.98 & 68.79 & 75.78 & -  \\
SSAN \cite{ding2021semantically} & 54.23 & 72.63 & 79.53 & -  \\ 
IVT \cite{ShuWWCSQRW22eccv} & 56.04 & 73.60 & 80.22 & - \\
LGUR \cite{ShaoZFLWD22mm} & 57.42 & 74.97 & 81.45 & - \\
\hline
\textbf{Ours} & \textbf{57.69} & \textbf{75.79} & \textbf{82.67} & \textbf{36.07}  \\
\hline
\end{tabular}
\end{table}

\subsection{Implementation Details}
Both textual and visual Transformers in our model have $L$ $=$ 12 layers, whose output features $\mathbf{f}_{T}$ and $\mathbf{f}_{I}$ have the same dimension $D$ $=$ 768. To get a better initial setup, we load weights of BERT \cite{devlin2018bert} as initial parameters of textual Transformer and weights of ViT \cite{dosovitskiy2021imageICLR} as initial parameters of visual Transformer.
The position embeddings in the textual Transformer keep the same setting as BERT. Since the resolution of input image is different from the original ViT implementation, the position embedding pretrained on ImageNet can not be directly loaded here. Therefore, we introduce interpolation to handle any given input resolution. Similar to ViT, the position embedding is also learnable.

As for the PDG module, a new text-image pair sample is generated for each person that satisfies the selection condition in the training dataset. During training, approximate text generation is randomly carried out with a probability of 0.5 for each sample in the batch. So is with feature-level mixup.  A text-image sample in the batch will be coupled with another sample randomly selected from the training dataset to fulfil mixup.

Following previous TBPS works \cite{ChenZLWZ22Neurocompt,Zhang_2018_ECCV}, we resize all input images to 384$\times$128 pixels. The size of image patch is 16. The training images are augmented with padding, random horizontal flipping, random cropping. All input text lengths are unified to 64 for training. The training data are enriched by the proximity data generation module. The batch size is set to 40 per GPU when training with Adam. The total number of epochs is set as 70. The base learning rate is initially set as $10^{-4}$, and decreased by 0.1 every 20 epochs. Besides, we initialize the learning rate by the warm-up in first 10 epochs. The default temperature $\tau$ of contrastive loss is set as 0.005. The parameters $\lambda$ and $\mu$ are set as 0.5 and 0.1, respectively. The stride step $w$ is 12.
All the experiments are performed with 4 Nvidia TeslaV100 GPUs.

In the testing phase, each image or sentence is forwarded only once to get its embedding feature, then we calculate their cosine similarity of them and get the recall images according to the query sentence.

\subsection{Comparisons with State-of-the-Art Methods}
We compare our model with existing state-of-the-art methods on two widely-used TBPS datasets CUHK-PEDES and ICFG-PEDES. 
CUHK-PEDES is slightly larger than ICFG-PEDES in scale.
We select the best-performing methods for comparison, and the results on the CUHK-PEDES dataset are reported in Table~\ref{tab:cuhk_sota}. Note that most recent methods either utilize multi-scale structures to match two modalities locally and globally \cite{ChenZLWZ22Neurocompt,farooq2021axm,gao2021contextual,niu2020improving} or take advantage of side information as auxiliary tasks \cite{jing2020pose,aggarwal2020text,wang2020vitaa} to help cross-modal representation learning. These works generally use CNN as backbone to extract image features and use LSTM or BERT (Transformer) as backbone to extract text features.
Compared to their sophisticated designs, we use pure Transformer architecture without any extra design for multi-scale and side information, and our elegant model outperforms the best-performing method IVT \cite{ShuWWCSQRW22eccv} by a significant margin, i.e., 3.88\%, 4.02\%, 2.92\% in terms of Top1, Top5, Top10 accuracy. The improvements are impressive compared to IVT's improvements from other methods.
It is worth noting that TextReID \cite{HanHZX21bmvc} exploits the CLIP model pretrained on large-scale vision-language databases to address the TBPS task, but it achieves inferior performance.
As for the ICFG-PEDES dataset, it has more images for each person but the total amount of identities is smaller, and there is only one sentence caption for an image, making less diverse in the cross-modal data. Table~\ref{tab:icfg_sota} shows the comparison results, and our model outperforms the best-performing method LGUR \cite{ShaoZFLWD22mm} by 0.27\%, 0.82\%, 1.22\% for Top1, Top5, Top10 accuracy, respectively.

\subsection{Ablation Study}
We conduct comprehensive ablation study experiments on the CUHK-PEHDES dataset to provide in-depth analysis of how the important modules and parameters affect the performance of our model.

\label{tau}
\begin{table}[!t]
\centering
\caption{Quantitive analysis of loss function}
\label{tab:tau}
\begin{tabular}{|c|cccc|}
\hline
Loss & Top1 & Top5 & Top10 & mAP \\
\hline
\hline
CMPM Loss & 55.21 & 76.38 & 83.87 & 47.60\\
InfoNCE Loss($\tau$=0.2) & 36.87 & 60.12 & 70.12 & 34.39\\
InfoNCE Loss($\tau$=0.07) & 62.63 & 81.64 & 87.96 & 54.31 \\
InfoNCE Loss($\tau$=0.005) & \textbf{69.47} & \textbf{87.13} & \textbf{92.13} & \textbf{60.56} \\
InfoNCE Loss($\tau$=0.001) & 67.96 & 85.89 & 91.26 & 59.11 \\
\hline
\end{tabular}
\end{table}

\begin{table}[!t]
\centering
\caption{Ablation study of overlap slicing (OS)}
\label{tab:OS}
\begin{tabular}{|c@{\hspace{1.5ex}}|c@{\hspace{2.5ex}}c@{\hspace{2.5ex}}c@{\hspace{2.5ex}}c|}
\hline
OS & Top1 & Top5 & Top10 & mAP \\
\hline
\hline
 $\times$ & 68.14 & 85.81 & 91.35 & 59.60 \\
 $\surd$  & \textbf{69.47} & \textbf{87.13} & \textbf{92.13} & \textbf{60.56} \\
\hline
\end{tabular}
\end{table}

\begin{table}[!t]
\centering
\caption{Ablation study of the PDG module}
\label{tab:PDG}
\begin{tabular}{|c@{\hspace{1.5ex}}c@{\hspace{1.5ex}}c|c@{\hspace{2.5ex}}c@{\hspace{2.5ex}}c@{\hspace{2.5ex}}c|}
\hline
TextImgGen  & TextGen & Mixup & Top1 & Top5 & Top10 & mAP \\
\hline
\hline
$\times$ & $\times$  & $\times$  & 65.49 & 84.53 &90.10 & 56.58 \\
$\surd$ & $\times$   & $\times$  & 67.85 & 86.04 & 91.09 & 58.29   \\
$\times$ & $\surd$   & $\times$   & 67.52 & 85.34 & 90.76 & 57.69  \\
$\times$ & $\times$   & $\surd$  & 66.27 & 85.14 & 90.63 & 57.43 \\
$\surd$ & $\surd$   & $\times$  & 68.68 & 86.50 & 91.52 & 58.88 \\
$\surd$  & $\surd$   & $\surd$ & \textbf{69.47} & \textbf{87.13} & \textbf{92.13} & \textbf{60.56} \\
\hline
\end{tabular}
\end{table}

\begin{figure*}[!ht]
   \centering
   \includegraphics[width=0.9\linewidth]{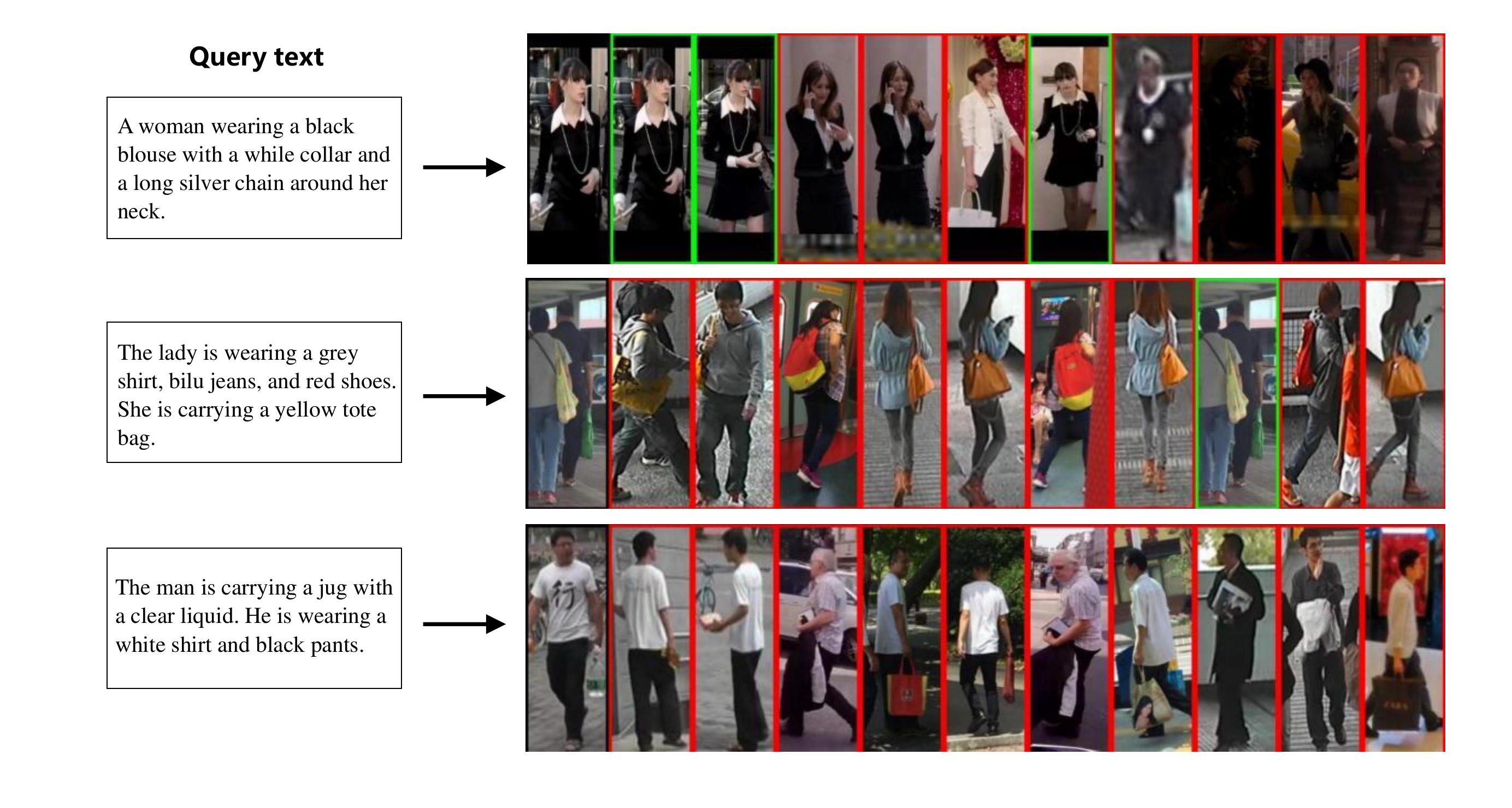}
   \caption{Visualization results of success and failure cases of our method. Given the query sentence, the first column is a ground truth person image and the latter ten columns are the Top10 images retrieved by our model. Green box denotes match while red one denotes mismatch.}
   \label{fig:vis}
\end{figure*}

\noindent\textbf{Effect of Loss Function:}
We implement the widely used loss for this task, i.e., CMPM \cite{Zhang_2018_ECCV}, on our model for comparison. As shown in Table \ref{tab:tau}, we found that, compared with CMPM, the InfoNCE Loss can let our model learn a better representation when the temperature is small enough. 

\noindent\textbf{Effect of Temperature Parameter:}
We show in Section \ref{sec:loss} that a small temperature $\tau$ plays a key role in utilizing the hardest samples for contrastive learning. In experiments, we found that the contrastive loss for our fine-grained model does be highly sensitive to the temperature $\tau$. Some representative performance results with respect to $\tau$ are reported in Table~\ref{tab:tau}.
We empirically change the parameter and observe that decreasing $\tau$ from 0.2 to 0.001, the Top1 accuracy can improve from 36.87\% to 69.47\%, which is consistent with our analysis presented in Section \ref{sec:loss}. As exhibited in Table~\ref{tab:tau}, we experimentally found $\tau$=0.005 is good enough for cross-modal representation learning of person search.

\noindent\textbf{Effect of Overlapping Slicing:}
The original patch slicing is non-overlapping in ViT, which may lose some informative feature, thus we propose to use overlapping slicing to get the input patches. As reported in Table~\ref{tab:OS}, overlapping slicing can improve the performance of our model at the cost of more computation.

\noindent\textbf{Effect of PDG Module:} 
The PDG module plays a crucial role in our method to address the data insufficiency problem. As described in Section \ref{sec:dataGeneration}, It includes three parts: controlled text-image pair generation (denoted as ``TextImgGen''), approximate text generation (denoted as ``TextGen''), and feature-level mixup (denoted as ``Mixup'').  Table~\ref{tab:PDG} reports the ablation study results on the PDG module. The first line shows the results of our model without the PDG module. It should be noted that, our simple dual Transformer model equipped with the hardness-aware contrastive learning strategy can outperform the state-of-the-art methods. Its Top1, Top5 and Top10 accuracy are 65.49\%, 84.53\% and 90.10\% respectively, while those of the best-performing method IVT are 65.59\%, 83.11\% and 89.21\%, as exhibited in Table~\ref{tab:cuhk_sota}. As revealed by the sixth line in Table~\ref{tab:PDG}, the PDG module further brings significant improvements to our model, increasing the  Top1, Top5 and Top10 accuracy by 3.96\%, 2.60\% and 2.03\%, respectively.

\begin{table}[!t]
\centering
\caption{Ablation study on controlled text-image pair generation (quantity for each person)}
\label{tab:textImgGen}
\begin{tabular}{|c|cccc|}
\hline
Quantity/ID & Top1 & Top5 & Top10  & mAP  \\ 
\hline
\hline
1 & 69.47 & 87.13 & 92.13 & 60.56 \\
2 & 69.56 & 86.96 & 92.02 & 60.59  \\
3 & 69.64 & 86.99 & 91.88 & 60.78  \\
4 & 69.79 & 86.78 & 91.83 & 60.41  \\
5 & 70.37 & 87.20 & 91.85 & 61.00  \\ 
6 & 70.13 & 86.99 & 91.75 & 60.85  \\ 
\hline
\end{tabular}
\end{table}

\begin{table}[!t]
\centering
\caption{Ablation study of approximate text generation}
\label{tab:textGen}
\begin{tabular}{|c@{\hspace{1.5ex}}c@{\hspace{1.5ex}}c|c@{\hspace{2.5ex}}c@{\hspace{2.5ex}}c@{\hspace{2.5ex}}c|}
\hline
SDEL & CDEL & REPL & Top1 & Top5 & Top10 & mAP \\
\hline
\hline
$\times$ & $\times$ & $\times$  & 64.45 & 83.33 & 89.31 & 55.67 \\
$\surd$ & $\times$  & $\times$   & 66.21 & 84.82 & 90.17 & 56.78   \\
$\times$ & $\surd$  & $\times$   & 66.01 & 84.71 & 90.72 & 56.65  \\
$\times$ & $\times$  & $\surd$   & 65.27 & 84.37 & 90.36 & 56.10   \\
$\surd$ & $\surd$   & $\surd$   & 66.34 & 85.12 & 90.75 & 56.81  \\
\hline
\end{tabular}
\end{table}

As shown in Table~\ref{tab:PDG}, among the three parts of the PDG module, the controlled text-image pair generation brings the most obvious improvement on the performance of our model. Currently, in order to make the automatic generation process under control, we only change the color of clothes to generate new text-image pairs. It is expected that more diverse samples can be generated to further improve the model performance when more controllable text-to-image generation models are developed.
As for approximate text generation, we can observe that it brings effective improvement to our model. Similar to controlled text-image pair generation, by providing approximate sentences to the original person image, it can implicitly enhance cross-modal alignment in model learning. 
Moreover, as shown in the fifth line of Table~\ref{tab:PDG}, by combing the two (i.e., TextImgGen and TextGen), the performance of the model is further enhanced, surpassing that of employing each of them alone, demonstrating their respective efficacy.
For feature-level mixup, as exhibited in Table~\ref{tab:PDG}, it can bring some extra improvement to our model by producing more diversified samples in the proximity feature space, but contribute the least among the three.

For controlled text-image pair generation, we conduct experiments to evaluate how the number of generated text-image pairs influences the model performance. We vary the number of generated text-image pairs for each person in the training net whose caption satisfies the selection condition.
The results are reported in Table \ref{tab:textImgGen}. As can be observed, when more generated text-image pairs of each person are used for training, the Top1 accuracy increases gradually from 69.47\% to 70.37\% and then decreases. On the other hand, minor fluctuations are observed in the Top5 accuracy, Top10 accuracy, and mAP. It indicates that generating many more text-image pairs does not necessarily lead to better model performance. This is because these data are generated in the proximity space of original data, and simply generating more data cannot consistently provide greater diversity.

In approximate text generation, three different text operations are exploited.  Therefore, we further conduct experiments to show how these operations affect our model's performance.
We solely employ each of them in approximate text generation to train our baseline dual Transformer model. 
The results are reported in Table~\ref{tab:textGen}.
As shown, all the three operations can bring extra gain, and  ``CDEL'' is the most effective one while ``REPL'' only brings a little improvement. Intuitively, ``CDEL'' deletes consecutive words (phases), which can improve the variety with weaker harming of the original context information, while ``SDEL'' randomly deletes words, which may change the semantic context and make it not as effective as ``CDEL''. The ``REPL'' operation can bring variety but may introduce noise, making it less effective. By combining the three text operations, our model can achieve an obvious improvement of 1.89\%, 1.79\%, 1.44\%, and 1.14\% in terms of Top1, Top5, Top10, and mAP accuracy, respectively, over the baseline one without any text operation.

\subsection{Visualization Analysis}
We present some representative retrieval results to qualitatively analyze the proposed method. Figure~\ref{fig:vis} visualizes several Top10 retrieval results of success and failure cases of our model. 
As shown, since person search is a fine-grained retrieval task, the model needs to distinguish the target from many similar samples. The first column shows an image of the ground truth person, and the correct retrieved images are denoted with green boxes.
The first row shows a representative successful case, where our model successfully distinguishes the target image by ``the silver chain'' with a Top1 match. And within the Top10 retrieval results, all three images of the targeted person are retrieved. The second case just retrieves one Top10 match. However, it is worth noting that, some other retrieved images seem acceptable since they match the text though are from different identities. The last row is a failure case with no Top10 match. We note that it is very challenging, where the model retrieves several similar persons with white shirt, black pants, carrying something in the hand. It fails only due to the visual feature ``a jug with liquid'' is too difficult to recognize.

\section{Conclusion and Future Work}
In this work, we have presented a simple yet effective dual Transformer model to address the text-based person search task. By introducing a hardness-aware contrastive loss, our model can achieve state-of-the-art results. In contrast to most previous TBPS works, our model learns cross-modal alignment implicitly without any sophisticated architecture designs. Furthermore, we design a proximity data generation module to address the data insufficiency problem of this fine-grained task, which automatically generates more diversified text-image samples in the proximity space of the original ones for model learning. It exploits recently-developed text-to-image diffusion models to fulfil controlled text-image pair generation, which is further coupled with approximate text generation and feature-level mixup.
Extensive experiments show that the proximity data generation module improves our model by significant margins. Since annotation is difficult and expensive for many tasks and real-world applications, we hope that our work can encourage more research to exploit the rapidly-developing generative AI methods to handle the data insufficiency problem of such tasks.

There are also some limitations should be addressed in our work. While we have employed generative models to produce new data, the quality of these generated data still requires improvement. This is primarily because existing generative models exhibit instability. In our efforts to generate acceptable data for training, we have imposed constraints on the range of generated data (i.e., only altering clothing colors in text-image pairs). Consequently, the diversity of the generated data remains limited. In future work, we intend to explore more advanced and effective generative models and develop better controlled auto-generation pipelines that yield a higher diversity of generated data. Additionally, we will investigate more complex text-based person search models and analyze the impact of generated data on their performance.


\bibliographystyle{IEEEtran}
\bibliography{reference}

\end{document}